\documentclass[english]{cccconf}
\usepackage{amsmath,amssymb}
\usepackage{amsthm}
\usepackage{psfrag}
\usepackage{epsfig}
\usepackage{graphicx,subfigure}
\usepackage{mathrsfs}
\usepackage{url}
\usepackage{bbm,color}
\usepackage[compress]{cite}
\usepackage{enumerate}
\usepackage{mathtools}
\usepackage{array}
\usepackage{supertabular}
\usepackage{psfrag,epsfig,graphicx,subfigure}
\usepackage{slashbox,multirow}
\graphicspath{ {images/} }

\begin{document}

\title{Incorporating Depth into both CNN and CRF for Indoor Semantic Segmentation}

\author{Jindong Jiang,
        Zhijun Zhang$^*$, \emph{Member, IEEE},
        Yongqian Huang,
        Lunan Zheng
        }

\affiliation{The School of Automation Science and Engineering, South China University of Technology\\ Guangzhou 510640, China\email{jdpshq@gmail.com, auzjzhang@scut.edu.cn, yongqianhuanggz@gmail.com, aulnzheng@sina.com}}

\maketitle

%%%%%%%%%%%%%%%%%%%%%%%%%%%%%%%%%%%%%%%%%%%%%%%%%%%%%%%%%%%%%%%%%%%%%%%%%%%%%%%%
\begin{abstract}

In this paper, we address the problem of indoor semantic segmentation by incorporating the depth information into the convolutional neural network and conditional random field of a neural network architecture. The architecture combines a RGB-D fully convolutional neural network (DFCN) with a depth-sensitive fully-connected conditional random field (DCRF). In the DFCN module, the depth information is incorporated into the early layers using a fusion structure which is followed by several dilated convolution layers for contextual reasoning. Later in the DCRF module, a depth-sensitive fully-connected conditional random field (DCRF) is proposed and combined with the previous DFCN output to refine the preliminary result. Comparative experiments show that the proposed DFCN-DCRF architecture achieves competitive performance compared with state-of-the-art methods.
\end{abstract}

\keywords{Convolutional neural networks, conditional random fields, RGB-D, semantic segmentation, transfer learning.}

%%%%%%%%%%%%%%%%%%%%%%%%%%%%%%%%%%%%%%%%%%%%%%%%%%%%%%%%%%%%%%%%%%%%%%%%%%%%%%%%
\section{Introduction}\label{sec.Introduction}

In order to realize scene understanding, semantic segmentation plays a very important role and has attracted more and more researchers’ interests \cite{he2004multiscale, shotton2009textonboost, kohli2009robust, koltun2011efficient, long2015fully}. Among existing methods, convolutional neural networks (CNNs) have shown great advantages on semantic segmentation with RGB images. One typical CNN, called fully convolutional neural network (FCN), achieves remarkable performance over the past few years.
As reported in Ref. \cite{long2015fully,noh2015learning,badrinarayanan2015segnet}, encoder-decoder type FCNs dramatically improved the dense prediction accuracy by fusing different layer representations. In order to expand the receptive field without losing resolution and generate a better performance on multiple segmentation tasks, a dilated convolution operator was applied to replace the encoder-decoder architecture \cite{yu2015multi,chen2016deeplab}. Despite many efforts had been taken on the improvement, the result was still unsatisfactory, especially, on the boundary of the objects. To remedy this problem, researchers started to combine RGB model based fully-connected conditional random fields (CRFs) with CNN and gained improvements on several semantic segmentation benchmarks \cite{yu2015multi,chen2016deeplab,lin2016exploring,schwing2015fully,zheng2015conditional,dai2015boxsup}. However, it is difficult to apply these methods in the indoor scene where objects share similar colors.

Recently, some RGB-D image datasets \cite{silberman2012indoor,janoch2011category,xiao2013sun3d,song2015sun} have been released in public. Since the depth information includes 3D positions and structures of the objects, utilizing depth channel as complementary information to RGB channel may increase the potential to implement accurate semantic segmentation. This hypothesis is verified by Couprie et al. who interacted the depth information into a multiscale convolutional network \cite{couprie2013indoor}. Inspired by this work, a novel neural network (DFCN) architecture with a depth-sensitive fully-connected conditional random field (DCRF) is proposed in this paper. Different from the existing FCNs, we incorporate the depth information into a FCN with dilated operator and a CRF to improve the accuracy greatly.

Before ending this section, the main contributions of this work are listed as follow:
\begin{enumerate}
\item[1.] A novel neural network architecture (termed DFCN-DCRF) is proposed, which combines an RGB-D fully convolutional neural network (DFCN) with a depth-sensitive fully-connected conditional random field (DCRF).
\item[2.] The design process and theoretical analysis of the proposed DFCN-DCRF is presented in detail.
\item[3.] Two comparison experiments on SUN RGB-D benchmark verify the effectiveness of the proposed DFCN-DCRF on semantic segmentation.
\end{enumerate}

\section{Related Work}
In this section, the literature of deep CNN for semantic segmentation, fully-connected conditional random fields, and incorporation of depth information are previewed in detail.

\subsection{Deep Convolutional Neural Network for Semantic Segmentation}

In 2015, Long et al. proposed a fully convolutional neural network model \cite{long2015fully}, which had a structure of encoder-decoder architecture. In this work, a skip architecture was designed, which combined semantic information from the deep coarse layer with appearance information from a shallow fine layer. The skip architecture is able to take advantage of all feature spectra and showed an accurate segmentation result. As a further discussion, Noh et al.  \cite{noh2015learning} proposed a novel FCN structure which eliminates the limitation of fixed-size receptive field. On the decoding step, it applied unpooling and convolution transpose to allow the network to learn the upsample weights. With the similar network structure of these two models, Badrinarayanan et al. \cite{badrinarayanan2015segnet} presented another architecture called SegNet, which comprised unpooling as well as the skip architecture. Besides, dropout \cite{srivastava2014dropout} and batch normalization \cite{ioffe2015batch} can also further improve the segmentation accuracy during test time \cite{badrinarayanan2015segnet,paszke2016enet}.

\begin{figure}[!t]
    \centering
    \subfigure[normal convolution]{
    \includegraphics[width=0.4\textwidth]{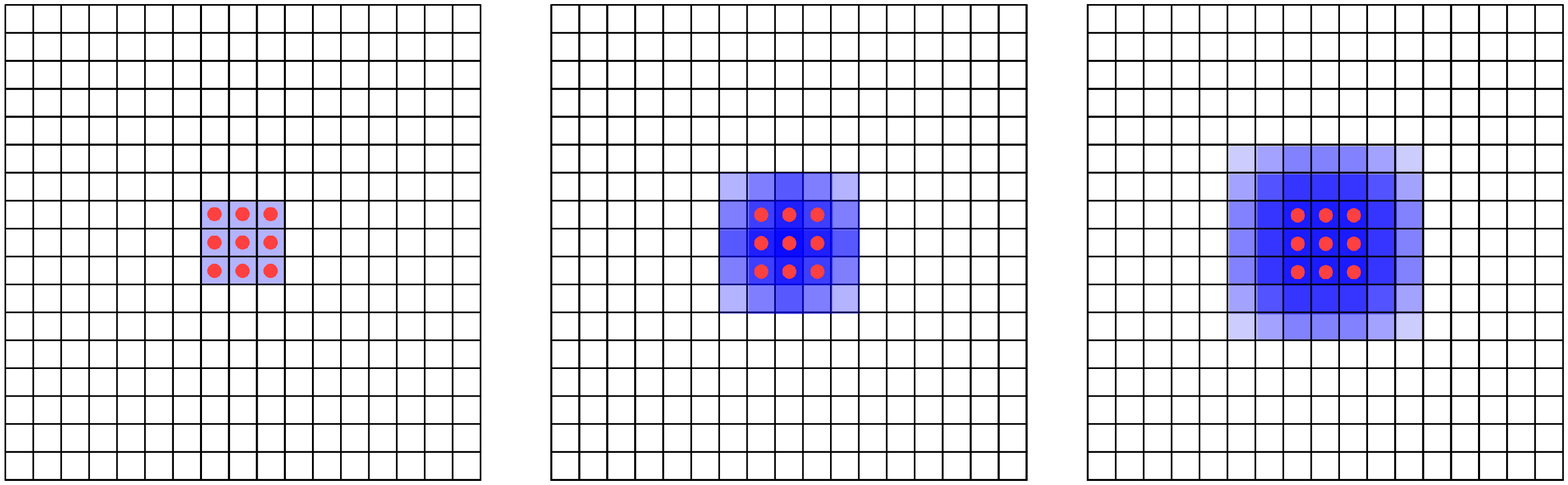}
    \label{fig:NormalConv}}
    \subfigure[dilated convolution]{
    \includegraphics[width=0.4\textwidth]{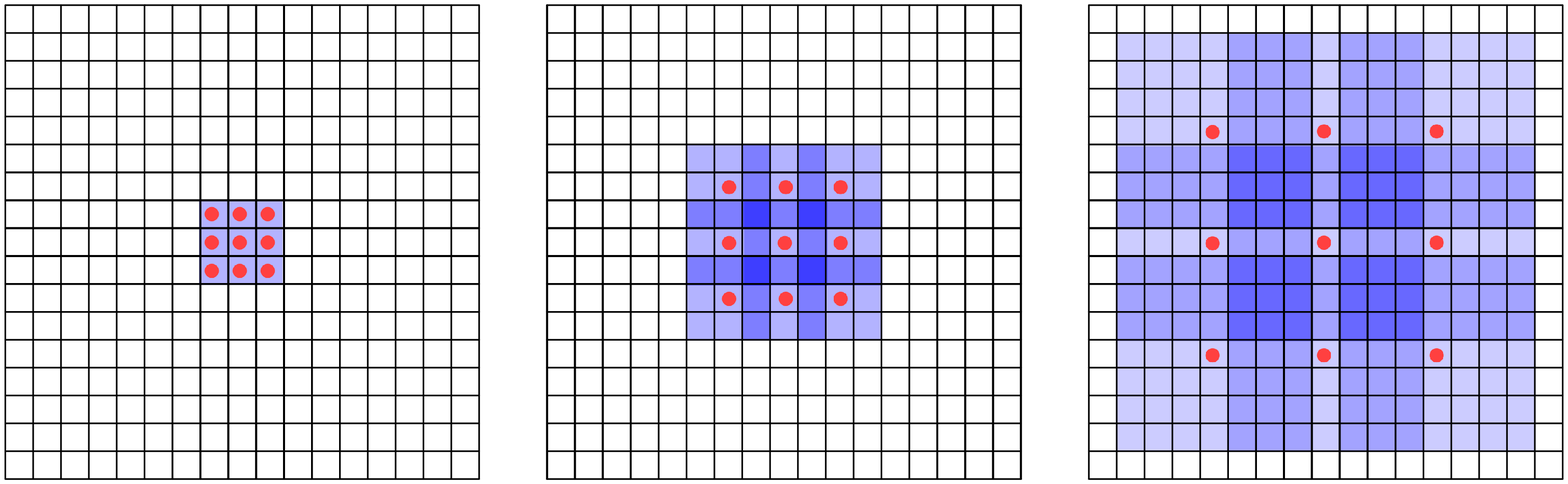}
    \label{fig:DilatedConv}}
    \caption{Receptive field of normal convolution and dilated convolution. Left to Right: the grids (marked in blue) contributes to the calculation of the center grids (marked in red) through three convolution layers with a kernel size of \(3\times3\). (a) Receptive field of normal convolution through three layers. (b) Receptive field of convolution layers with 1, 2, and 4 dilation rate through three layers.}
    \label{fig:Convolution}
\end{figure}

Meanwhile, another approach for contextual reasoning called dilated convolution was proposed. It aims to expands the receptive field on the input image exponentially without losing resolution of the result. As shown in Fig. \ref{fig:Convolution}, dilation on convolution expands the receptive field on the input image exponentially without losing resolution of the result. Based on this architecture, Chen et al. proposed DeepLab system which conducts contextual reasoning on images while keeping the spatial information \cite{chen2016deeplab}. Yu et al. further applied dilated convolution on a novel FCN architecture with no pooling layer  \cite{yu2015multi}. In our work, we design the core network with both pooling and dilated convolution. The fundamental structure of the proposed DFCN-DCRF is similar to the DeepLab-LargeFOV architecture \cite{chen2016deeplab}.

\subsection{Fully-Connected Conditional Random Fields}

Recently, some semantic segmentation algorithms based on CNN combine the FCN with conditional random fields (CRFs). CRFs is able to model the contextual relationships between different pixel so as to maximize the label agreement of them. Koltun et al presented an efficient inference algorithm for Gaussian Edge Potentials \cite{koltun2011efficient}. The inference method allows a fully-connected CRF with pairwise connection over all pairs of pixels to inference in a reasonable time. It has been proved that the poor accuracy of boundary in the output of FCN can be addressed by combining the responses in the last layer of CNN with a fully-connected CRF model  \cite{chen2016deeplab,yu2015multi,lin2016exploring}.
In particular, Zheng et al. proposed a novel architecture, in which the mean field approximation was modeled as a recurrent neural network and integrated as a part of deep neural network \cite{zheng2015conditional}.

Fully-connected CRFs with RGB information works well on refining CNN output, based on the fact that different objects have different colors or brightness. However, indoor scene objects (e.g., bed, couch, pillow) often share similar color or brightness. Therefore, it is reasonable to incorporate the depth information into fully-connected CRF as a post-processing method to provide additional information such as distance or clear distinctive boundaries. The idea of incorporating depth information into conditional random fields was first proposed by Muller et al. \cite{muller2014learning}. They applied a super-pixel-based model for semantic segmentation. Inspired by Ref. \cite{muller2014learning}, we incorporate the depth information into a fully-connected CRF after CNN architecture to generate a more accurate segmentation.

\subsection{Incorporation of Depth Information}

Based on some labeled RGB-D image datasets \cite{silberman2012indoor,janoch2011category,xiao2013sun3d,song2015sun}, many studies have tried to incorporate the depth information to generate a better performance. Long et al. \cite{long2015fully} shows that simply stack the depth with RGB as a 4-channels input cannot improve the performance in a significant way. Gupta et al. \cite{gupta2014learning} presented a different representation of depth information referred as HHA. It comprised horizontal disparity, height from the ground and the angle between the local surface normal and gravity direction, and got a good results \cite{long2015fully,gupta2014learning}. However, Hazirbas et al. \cite{hazirbas2016fusenet}.  argued that the HHA representation did not contain more information than the raw depth itself, and it required a high computation cost. In their work, a fusion-based CNN architecture was presented. The network consisted of two branches of encoding networks, i.e., a depth branch and an RGB branch. The feature representation in these two branches was then fused into the master branch. There are two ways of fusing approaches, i.e., sparse fusion and dense fusion. It was proved that the sparse one is better. Therefore, in our work, we fuse RGB and depth channel feature representation in a sparse way from Conv1 to Conv4.

\section{Approach}

In this section, the design process and theoretical analysis of the proposed DFCN-DCRF is stated in detail.

\begin{figure*}[t]
    \centering
    \includegraphics[width=0.75\textwidth]{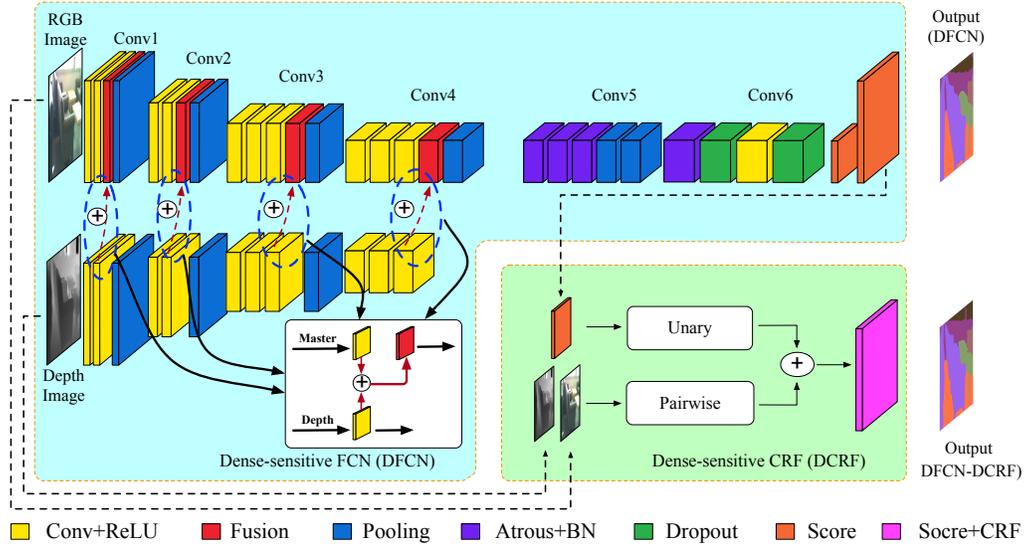}
    \caption{The proposed DFCN-DCRF architecture.}
    \label{fig:Architecture}
\end{figure*}

\subsection{RGB-D FCN for Unary Potential Generation}\label{CNN}

We propose a RGB-D FCN architecture (DFCN) to generate the response for the unary potential for each class on each pixel. As shown in Fig. \ref{fig:Architecture}, the DFCN part has two major blocks: 1) Convolution layers with three downsample pooling for features extraction and depth fusion; 2) Dilated convolution layer for contextual reasoning and dense prediction. 

In the first block, we employ the 16-layer VGG net from first layer Conv1\_1 to Conv4\_3 as a fundamental framework. This fundamental framework is applied on both the RGB channel and depth channel (i.e., master branch and depth branch) for features extraction. In this stage, we take layers before every pooling in two branches of the network and fuse them together with element-wise summation. The fusion layers are then added before every pooling layer in the master branch. To prevent further resolution decrease of the feature maps, we replace Pool4 in VGG-16 with a one-strike max-pooling and Pool5 with a one-strike max-pooling and a one-strike average-pooling. To make the values in two branches compatible and easier to train, we normalize the depth channel, which originally ranges from 0 to 65535, into the same range as color images, i.e., from 0 to 255.

In the second block, the dilated convolution is applied after Pool4 with dilation rate of 2 for three layers in Conv5 and dilation rate of 12 for  Conv6\_1. All dilated convnets in the proposed DFCN-DCRF architecture are followed by batch normalization layers to avoid covariate shift \cite{ioffe2015batch}. Conv6\_1 and Conv6\_2 are also followed by dropout layers when training to avoid overfitting \cite{srivastava2014dropout}. The final score map is upsampled with a factor of 8 by using bilinear interpolation to recover the original resolution. This score map is also converted into a preliminary pixel-wise label prediction, that is, the result of DCRF.

\subsection{Depth-Sensitive Fully-Connected Conditional Random Fields}

We present a depth-sensitive fully-connected CRF (DCRF) to refine the upsampled output from DFCN. Every pixel is treated as a CRF node, and the energy function of the DCRF is composed of a unary and a pairwise factors (also called first and second order factors). Considering an image \textbf{I} has a size of $N$, the energy function $E(\mathbf{y})$ with $\mathbf{y}$ denoting label vector is defined as
\begin{equation}
  E(\mathbf{y})=\sum_i\phi_i(y_i) +
  \sum_{ij}\phi_{ij}(y_i, y_j)
\end{equation}
where $\mathbf{y}=[y_1,y_2,\cdots,y_i,\cdots,y_N]^\text{T}$ with $i\in[1,N]$, superscript $^\text{T}$ denoting the transpose of a matrix or a vector. Element $y_i$ is the label assigned of the $i$th pixel. The unary potential \(\phi_i(y_i)=-\log{P(y_i)}\) is computed from the last layer of DFCN, where \(P(y_i)\) is the result of applying softmax on the score map at pixel \(i\). \(\phi_{ij}(y_i, y_j)\) is a pairwise potential function with Gaussian kernel over all pair of pixels in image \textbf{I}, which can be represented as
\begin{equation}\label{eq:pairwise}
	\phi_{ij}(y_i, y_j)=
	\mu(y_i, y_j)\big[\omega_1\theta_a(\mathbf{f}_i, \mathbf{f}_j) + \omega_2\theta_s(\mathbf{f}_i, \mathbf{f}_j)\big]
	\
\end{equation}
where $\mu$ is the label compatibility function. In our model, \(\mu(y_i, y_j)=[y_i \neq y_j]\). In Potts model \cite{kohli2007p3}, it means that we have penalty for the assignment of different labels. \(\mathbf{f}_i\) and \(\mathbf{f}_j\) are feature vectors of pixels in the $i$th and $j$th positions. \(\theta_s(\mathbf{f}_i, \mathbf{f}_j)\) is smoothness kernel, i.e.,
\begin{equation}
    \theta_s(\mathbf{f}_i, \mathbf{f}_j)=
	\exp\bigg(
    	-\frac{\parallel p_{i}-p_{j} \parallel^2}
        {2\sigma_{\gamma}^{2}}
    \bigg)
\end{equation}
where \(p_i\) and \(p_j\) denote the position vectors of the $i$th and $j$th pixels. Parameter $\sigma_\gamma$ controls the degrees of nearness of two pixels. The smoothness kernel is used to eliminate small isolated regions. \(\theta_a(\mathbf{f}_i, \mathbf{f}_j)\) is appearance kernel. In this paper, we present two kinds of appearance kernels, i.e.,
\begin{equation}\label{eq:best}
\begin{split}
	\theta_a(\mathbf{f}_i, \mathbf{f}_j)=
	\exp\bigg(
    	-\frac{\parallel p_i-p_j \parallel^2}
        {2\sigma_{\alpha}^{2}}
        -\frac{\parallel I_i-I_j \parallel^2}
        {2\sigma_{\beta}^{2}} \\
	    \qquad \qquad {} -\frac{\parallel d_i-d_j \parallel^2}
        {2\sigma_{\nu}^{2}}
    \bigg)\\[10pt]
\end{split}
\end{equation}
%\textbf{or}
\begin{equation}\label{eq:alt}
\begin{split}
    \theta_a(\mathbf{f}_i, \mathbf{f}_j)=
	\exp\bigg(
    	-\frac{\parallel p_i-p_j \parallel^2}
        {2\sigma_{\alpha}^{2}}
        -\frac{\parallel I_i-I_j \parallel^2}
        {2\sigma_{\beta}^{2}}
    \bigg) \\
    +\lambda \exp\bigg(
    	-\frac{\parallel p_i-p_j \parallel^2}
        {2\sigma_{\alpha}^{2}}
	     -\frac{\parallel d_i-d_j \parallel^2}
        {2\sigma_{\nu}^{2}}
    \bigg)
\end{split}
\end{equation}
where \(p_i\) is defined the same as before, \(I_i\) is the color vector of the $i$th pixel and \(d_i\) is the depth vector of the $i$th pixel. $\sigma_{\alpha}$, $\sigma_{\beta}$, and $\sigma_{\nu}$ control the degrees of nearness and similarity between two pixels. With this definition, pixels with close position, similar color and similar depth are forced as the same label. The position, color and depth features are combined into one Gaussian kernel in Equation \eqref{eq:best} but two Gaussian kernels in Equation \eqref{eq:alt}, where \(\lambda\) controls the balance between two kernels. Equation \eqref{eq:best} indicates that big differences in either RGB channel or depth channel can cause the different assignments of labels, and thus the penalty will be small. On the contrary, Equation \eqref{eq:alt} only gives a small penalty for pixels whose RGB information and depth information are alike. In practice, we find that Equation \eqref{eq:best} provides better performance than Equation \eqref{eq:alt} in the context of indoor semantic segmentation.

In order to balance the importance of the depth and RGB information, the depth input to the fully-connected CRF must be scaled into a compatible range referred to RGB channel. The most accessible way to do so is to directly scale the depth channel into the range of RGB channel. However, the depth channel contains invalid values, which is always presented as an extreme value, i.e., 0 or 65535. These invalid values might prevent the scaled depth values from falling into an appropriate range. Therefore, rather than rigidly scale the depth into 0 to 255, we scale and shift every depth image to have the same mean value and standard deviation with its RGB counterpart. This allows the depth image and RGB image have compatible value range in CRF model.

\section{Experimental Evaluation}

In this section, the proposed DFCN-DCRF is tested on a SUN RGB-D scene understanding benchmark suite \cite{song2015sun}. This dataset was captured by four different kinds of sensors with different resolutions and fields of view. It also contains the data from NYU Depth v2 \cite{silberman2012indoor}, Berkeley B3DO \cite{janoch2011category} and SUN3D \cite{xiao2013sun3d} with totally 10,335 RGB-D images and their pixel-wise semantic annotations. Moreover, it has a default trainval-test split which comprises 5285 images for training/validation and 5050 images for testing. To improve the quality of depth channel, multiple frames are collected to obtain a refined depth map. However, we find that if the invalid area in the raw depth map is too large, the corresponding refined depth image still contains invalid measurement or losing information on corresponding pixels. Thus 387 training images are excluded, as they have more than \(45\%\) of invalid values in the raw depth map. According to Ref. \cite{eigen2015predicting}, since different classes of objects have different instance-wise and pixel-wise present frequency, we also use weighted losses for different classes.

\textbf{Training}~
For the CNN stage, bilinear interpolation on RGB channel and nearest-neighbor interpolation on depth channel are applied to get \(480\times480\)-size image inputs of two branches. The loss function is the sum of softmax loss on each pixel in the output map. The parameters before Pool4 are initialized with the values from the 16-layer VGG model \cite{simonyan2014very} pre-trained on ImageNet dataset \cite{russakovsky2015imagenet}. Since Conv1\_1 layer in depth branch only has one channel, we average the parameter values of VGG Conv1\_1 along 3 channels to get a single channel for its initialization. During training, the data is augmented by applying random hue, brightness, contrast and saturation adjustment on the original image, and we randomly scale and crop the image as well as the label to generate more data.

The proposed DFCN-DCRF architecture is implemented on the TensorFlow framework \cite{abadi2016tensorflow}, and stochastic gradient descent (SGD) is applied for end-to-end training. We set the initial learning rate of layers before Pool5 in the master branch as 0.0002, the final score layer as 0.005, and all other layers as 0.001. All those learning rates are decayed by a factor of 0.9 in every 50,000 iterations. A momentum of 0.9 and weight decay of 0.0005 are also applied. The network is continually trained on a Nvidia Titan X Pascal GPU with a batch size of 5 until the loss does not further decrease.

For the fully-connected CRF stage, we first obtain the DFCN response on the score layer after it is fine-tuned on training. \(\omega_{2}\) in Equation \eqref{eq:pairwise} is set as 3 and \(\sigma_{\gamma}\) in Equation \eqref{eq:best} is set as 3. Then a random search algorithm is employed to determine the best values for \(\omega_1\), \(\sigma_{\alpha}\), \(\sigma_{\beta}\) and \(\sigma_{\nu}\). More concretely, we randomly search the best values of \(\omega_1\) in a range from 5 to 11, \(\sigma_{\alpha}\) in a range from 90 to 170,  \(\sigma_{\beta}\) and \(\sigma_{\nu}\) in a range from 7 to 12 , which iteratively refines the search step around the last round's best values.

\textbf{Testing}~
The network is performed on 5050 images testing set with three criteria, i.e., the pixel accuracy, the mean accuracy and the intersection-over-union (IoU) score. \(C_{ij}\) denotes the number of pixels those are predicted as category \(j\) but actually belongs to category \(i\). \(C_{ii}\) denotes the number of pixels with correct prediction of category \(i\). \(T_i\) denotes the total number of pixels that belongs to category \(i\) in the ground truth. \(K\) denotes the total number of categories in the dataset.
\begin{enumerate}[i)]
    \item Pixel accuracy measures the percentage of correctly classified pixels:
    \begin{equation*}
        \text{Pixel} = \frac{\sum_{i}C_{ii}}{\sum_{i}T_{i}}
    \end{equation*}
    \item Mean accuracy measures the classwise pixel accuracy:
    \begin{equation*}
        \text{Mean} = \frac{1}{K}\sum_{i}\frac{C_{ii}}{T_{i}}
    \end{equation*}
    \item Intersection-over-union calculates the average value of the intersection between the ground truth and the prediction regions:
    \begin{equation*}
        \text{IoU} = \frac{1}{K}\sum_{i}\frac{C_{ii}}{T_{i} + \sum_{j}C_{ij} - C_{ii}}
    \end{equation*}
\end{enumerate}

Among these metrics, Pixel accuracy measurement is more sensitive to the large objects such as bed, wall, and floor in the dataset, so Pixel accuracy measurement will be misleading when the network performs better on the large objects. Therefore, Mean accuracy measurement and IoU score measurement are more informative.

\subsection{Quantitative Results}

Two comparison experiments are conducted in this section. In the first experiment, the proposed DFCN and DFCN-DCRF are compared with the start-of-the-art methods. The results are shown in Table \ref{tab:compare_sota}. For comparisons, we also illustrate the results of pure DFCN. The segmentation results show that both the DFCN and DFCN-DCRF outperform other existing methods, except for Context-CRF \cite{lin2016exploring}. However, it is worth pointing out that Context-CRF requires CNNs with multi-scale image input and pyramid pooling to generate CRF potentials for primary results and another dense CRF to do refinement, which is relatively difficult to train and inference compared with our method. It is also worth noting that our CNN method, which has no CRF post-processing stage, already outperforms FuseNet, which shares the same fusing strategy on a encoder-decoder based FCN architecture.
\begin{table}[t]
\centering
\caption{Comparison of segmentation results among the proposed DFCN-DCRF, DFCN and the state-of-the-art on SUN RGB-D benchmark \cite{xiao2013sun3d}. }
\label{tab:compare_sota}
\centering
\begin{tabular}{ p{8em} m{4em} m{4em} m{4em}}
\hline
	  & Pixel & Mean & IoU \\ \hline
	FCN-32s \cite{long2015fully} & 68.35 & 41.13 & 29.00 \\
	FCN-16s \cite{long2015fully} & 67.51 & 38.65 & 27.15 \\
	SegNet \cite{badrinarayanan2015segnet} & 71.2 & 45.9 & 30.7 \\
	Context-CRF \cite{lin2016exploring} & 78.4 & 53.4 & 42.3 \\
	FuseNet-SF5 \cite{hazirbas2016fusenet} & 76.27 & 48.3 & 37.29 \\
	DFCN & 76.1 & 50.8 & 38.0 \\
	DFCN-DCRF & 76.6 & 50.6 & 39.3 \\ \hline
\end{tabular}
\end{table}
\begin{table}[t]
\centering
\caption{Segmentation results of different depth incorporation strategies.}
\label{tab:self_compare}
\centering
\begin{tabular}{p{11em} m{3em} m{3em} m{3em} }
\hline
	  & Pixel & Mean & IoU \\ \hline
	DFCN$_\text{noDepth}$  & 72.4 & 44.7 & 33.4 \\
	DFCN$_\text{noDepth}$-CRF$_\text{noDepth}$ & 73.8 & 44.1 & 34.4 \\
	DFCN$_\text{noDepth}$-DCRF & 73.5 & 44.2 & 34.4 \\
	DFCN & 76.1 & 50.8 & 38.0 \\
	DFCN-CRF$_\text{noDepth}$ & 76.2 & 48.7 & 38.5 \\
	DFCN-DCRF & 76.6 & 50.6 & 39.3 \\ \hline
\end{tabular}
\end{table}

\begin{figure}[t]
    \centering
    \includegraphics[width=0.85\linewidth]{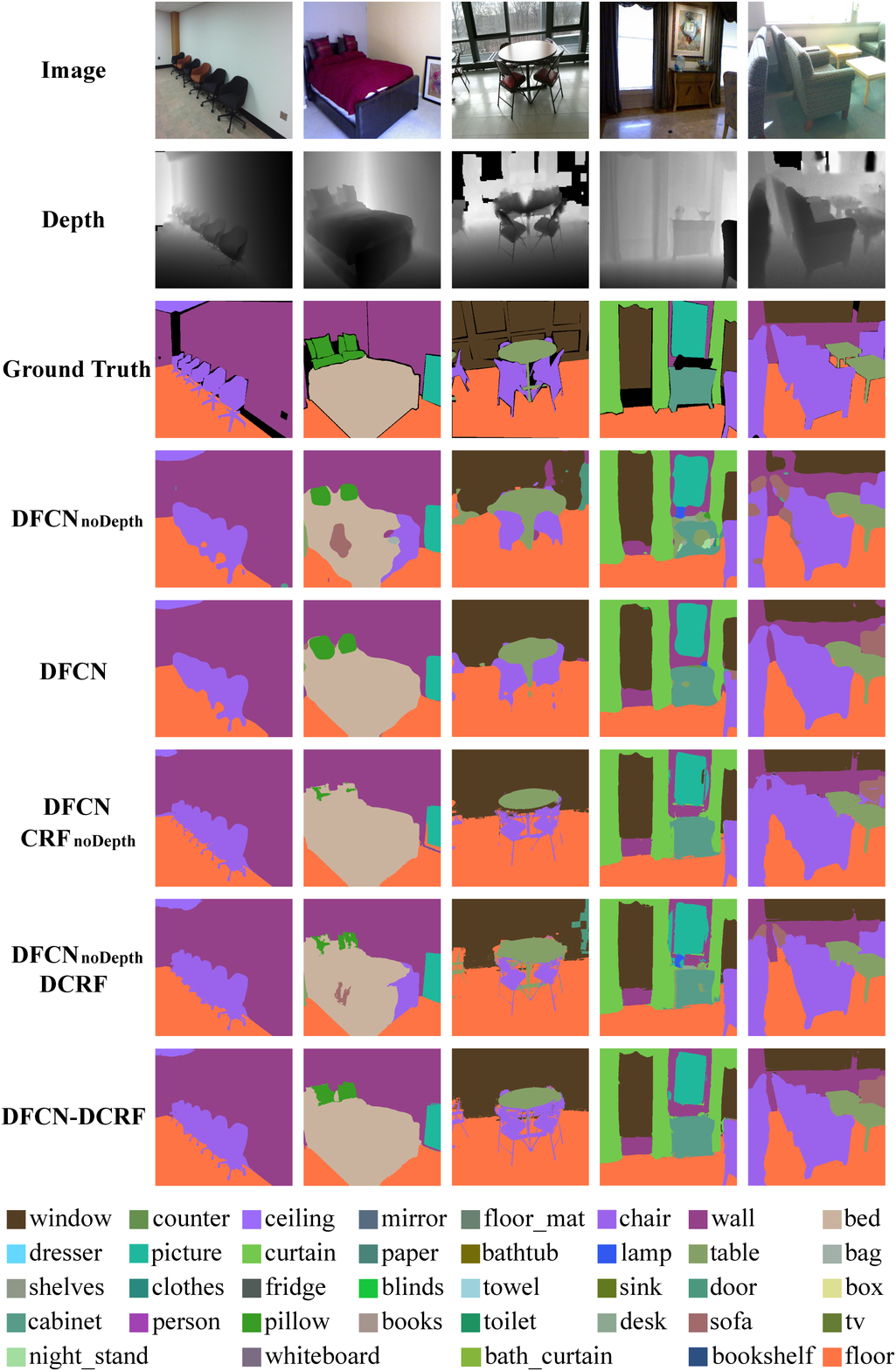}
    \caption{Visualization result of different depth incorporation cases on SUN RGB-D testing data.}
    \label{fig:image_comparison}
\end{figure}

In the second experiment, we test the proposed DFCN-DCRF with different depth incorporation cases, and the results are shown in Table \ref{tab:self_compare}. The CRF with and without depth are denoted as DCRF and CRF$_\text{noDepth}$. It can be seen from Row 2 and Row 3 of Table \ref{tab:self_compare} that if the RGB based CRF is integrated into RGB based DFCN, the performance will become a bit better but not much. That is to say, the RGB based CRF after RGB based DFCN can only cause a little improvement in Pixel accuracy measurement and IoU score measurement, and even obtain a setback in Mean accuracy measurement. The possible reason is that objects in the indoor environment often have similar colors. Therefore an RGB based CRF cannot distinguish the differences between the objects with similar colors. In addition, we add the depth information into only DCRF part of the model as shown in Row 4 of Table \ref{tab:self_compare}, the performance is further improved a bit. The result of adding depth information into DFCN is shown in Row 5 of Table \ref{tab:self_compare}, which shows that the performance increase a lot compared with the previous three cases. It implies that the depth information has a great effect if it is added into the DFCN. Furthermore, we combine the DFCN with the RGB based CRF, we find that the performance improves a bit but still not much as shown in Row 6 of Table \ref{tab:self_compare}. Compared with Row 2 and 3, as well as 5 and 6, whether integrating the RGB-based CRF into DFCN has little influence on the performance. As can be seen, the proposed DFCN-DCRF has the best performance among all cases. The same conclusion can be obtained from Fig. \ref{fig:image_comparison}. Moreover, we also compared our method with one state-of-the-art method, i.e., FuseNet-SF5 in classwise mIoU score. The results are shown in Table \ref{tab:classwise_compare}. It shows that both the proposed DFCN-DCRF and DFCN are better than FuseNet-SF5 Ref. \cite{hazirbas2016fusenet}.
\begin{table*}[t]
\scriptsize
\centering
\caption{mIoU score classwise comparison of FuseNet-SF5 in \cite{hazirbas2016fusenet}, our DFCN, and DFCN-DCRF}
\label{tab:classwise_compare}
\begin{tabular}{|l|c|c|c|c|c|c|c|c|c|c|c|c|c|}
\hline
	& wall & floor & cabin & bed & chair & sofa & table & door & wdw  & bslf  & pic  & cnter  & blinds \\ \hline
	SF5 \cite{hazirbas2016fusenet} & \textbf{74.94} & \textbf{87.41} & 41.70 & \textbf{66.53} & 64.45 & 50.36 & \textbf{49.01} & \textbf{33.35} & 44.77 & 28.12 & 46.84 & 27.73 & 31.47 \\
	DFCN & 74.72 & \textbf{87.41} & 41.52 & 62.49 & 64.58 & 48.78 & 44.94 & 31.54 & 46.18 & 31.08 & \textbf{47.71} & 31.09 & 31.15 \\
	DFCN-DCRF & 74.29 & 86.78 & \textbf{43.44} & 64.25 & \textbf{64.80} & \textbf{51.6} & 45.73 & 31.67 & \textbf{47.64} & \textbf{32.55} & 46.43 & \textbf{32.00} & \textbf{32.07} \\ \hline
	  & desk & shelf  & ctn  & drssr  & pillow & mirror & mat  & clthes  & ceiling & books & fridge & tv & paper \\ \hline
	SF5 \cite{hazirbas2016fusenet} & 18.31 & \textbf{9.20} & 52.68 & 34.61 & \textbf{37.77} & \textbf{38.87} & 0 & 16.67 & \textbf{67.34} & 27.29 & 31.31 & 31.64 & 16.01 \\
	DFCN & 20.47 & 7.16 & 53.58 & 35.65 & 35.50 & 28.57 & 0 & 26.18 & 64.46 & \textbf{33.32} & 37.82 & 36.34 & 22.21 \\
	DFCN-DCRF & \textbf{21.28} & 7.23 & \textbf{55.5} & \textbf{39.49} & 34.41 & 28.55 & 0 & \textbf{28.64} & 63.11 & 33.12 & \textbf{42.33} & \textbf{42.96} & \textbf{23.03} \\ \hline
	  & towel & shwr  & box & board  & person & stand  & toilet & sink & lamp & btub  & bag & mean &   \\ \hline
	SF5 \cite{hazirbas2016fusenet} & 16.55 & \textbf{6.06} & 15.77 & \textbf{49.23} & 14.59 & \textbf{19.55} & 67.06 & \textbf{54.99} & \textbf{35.07} & \textbf{63.06} & 9.52 & 37.29 &   \\
	DFCN & 28.43 & 0.21 & 23.62 & 45.03 & 29.64 & 16.27 & 65.94 & 48.84 & 33.74 & 56.08 & 15.41 & 38.0 &   \\
	DFCN-DCRF & \textbf{29.77} & 0 & \textbf{25.69} & 45.21 & \textbf{35.14} & 18.52 & \textbf{67.72} & 49.91 & 33.24 & 60.64 & \textbf{16.52} & \textbf{39.3} & \  \\ \hline

\end{tabular}
\end{table*}

\section{Discussion}

In this paper, a novel neural network architecture (termed DFCN-DCRF) has been designed and  proposed, which combines an RGB-D fully convolutional neural network (DFCN) with a depth-sensitive fully-connected conditional random fields (DCRF). Different from most of methods only adding depth-information into FCN, we have added the depth information into both the DFCN and DCRF.
In addition, the design process and theoretical analysis of the proposed DFCN-DCRF have been presented in detail. Two comparison experiments on SUN RGB-D benchmark have verified the effectiveness and advantages of the proposed DFCN-DCRF on semantic segmentation.

%\printbibliography

% \begin{thebibliography}{99}
% \bibitem{long2015fully}
% Long, Jonathan, Evan Shelhamer, and Trevor Darrell. "Fully convolutional networks for semantic segmentation." In Proceedings of the IEEE Conference on Computer Vision and Pattern Recognition, pp. 3431-3440. 2015.

\bibliographystyle{ieeetr}
\bibliography{reference}

\end{document}